**+Autonomous Multi-Rotor UAVs: A Holistic Approach to Design, Optimization, and Fabrication**

[1]Aniruth A, [1]Chirag Satpathy, [1]Jothika K, [1]Nitteesh M, [1]Gokulraj M, [3]Venkatram K, [4]Harshith G, [5]Shristi S, [1]Anushka Vani, [1]Jonathan Spurgeon

National Institute of Technology, Tiruchirappalli,
near BHEL, NH-83, Thuvakkudi, Tamil Nadu, India 620015.

## ABSTRACT

Unmanned Aerial Vehicles (UAVs) have become pivotal in domains spanning military, agriculture, surveillance, and logistics, revolutionizing data collection and environmental interaction. With the advancement in drone technology, there is a compelling need to develop a holistic methodology for designing UAVs. This research focuses on establishing a procedure encompassing conceptual design, use of composite materials, weight optimization, stability analysis, avionics integration, advanced manufacturing, and incorporation of autonomous payload delivery through object detection models tailored to satisfy specific applications while maintaining cost efficiency. The study conducts a comparative assessment of potential composite materials and various quadcopter frame configurations. The novel features include a payload-dropping mechanism, a unibody arm fixture, and the utilization of carbon-fibre-balsa composites. A quadcopter is designed and analyzed using the proposed methodology, followed by its fabrication using additive manufacturing and vacuum bagging techniques. A computer vision-based deep learning model enables precise delivery of payloads by autonomously detecting targets. This comprehensive design methodology is validated through successful flight tests of the fabricated drone.

## 1    INTRODUCTION

The unique attribute of operating an Unmanned Aerial Vehicles, autonomously or manually enhances the efficiency of UAVs and enables them to undertake critical missions. The widespread potential of UAVs to revolutionize diverse sectors [1,2] is evident through their extensive applications in defence, agriculture, healthcare, publicity, logistics, passenger transport, and consumer markets. The increasing versatility of drone applications and rapid technological advancements have fuelled remarkable growth in commercial and recreational drone markets. Notably, 2020 witnessed a staggering worldwide shipment of 526,000 commercial drones, marking an incredible 50% surge from 2019[3]. The escalating adoption of additive manufacturing (AM) and using composite materials to produce Unmanned Aerial Vehicle (UAV) structures is an emerging trend. This predilection towards AM can be attributed to its exceptional capacity of fabricating intricate, lightweight geometries without requiring traditional manufacturing methods, including moulding for plastics and composites and machining, drilling, and lathe techniques for metals, are inherently labour-intensive, often restricting design flexibility. In contrast, additive manufacturing encompasses a diverse range of processes, encompassing selective laser sintering (SLS), stereolithography (SLA), and fused deposition modelling (FDM) for polymers and composites. Also, composite manufacturing techniques cater to a high and reliable strength-to-weight ratio. Each technique boasts distinct advantages and disadvantages, rendering them suitable for specific applications. With such advancements and various possible options, there is a need to devise a holistic design methodology to enable effective and faster design of UAVs for the required application by incorporating and exploiting the advantages of composite materials and advanced manufacturing techniques. This study aims to devise a generalized conceptual design methodology for a multirotor UAV, examining each development phase, from preliminary design to the intricacies of automation. This comprehensive resource, critically assessing the merits and potential ramifications of the processes, will prove invaluable to scholars, industry professionals, and all stakeholders vested in the design and evolution of UAVs.

## 2    DESIGN METHODOLOGY

A structured design methodology has been formulated through rigorous research, which comprises the following sequence. The design constraints and requirements of the UAV are thoroughly analyzed for the specific application. A comparative study of the various multirotor frame configurations, considering factors like structural integrity, compactness, ease of manufacturing, manufacturing cost, propeller clearance, and manoeuvrability, yields the best-suited one for the desired application. An initial weight estimate is performed. The subsequent step involves selecting an appropriate motor and propeller combination. After reviewing data sheets for various combinations based on the requirements, the best-suited option is chosen. Components like the electronic speed controller (ESCs) and the battery are chosen to match the performance and endurance of the motors. With electronic components determined, a preliminary design containing the frame, arms, etc., is ideated for encompassing them while allowing for effective heat dissipation through the assistance of an external airflow and unrestricted access to all the components of maintenance and repair. Material selection is critical, guided by yield strength, maximum deflection, the factor of safety, cost, environmental impact, and part-specific requirements. Local and global structural analysis and topology optimization use the finite element method to reduce the total weight and material usage. Centre of gravity and stability assessments are assessed to achieve optimal results. These steps undergo multiple iterations until all constraints are met. Finally, a prototype based on this conceptual design is built and rigorously tested through numerous flight tests to validate the design methodology. This approach is holistic, spanning from conceptualization to fabrication. Leveraging advanced materials like carbon-fiber-balsa sandwich composite structures allows us to achieve an optimal strength-to-weight ratio. The UAV features unique elements, such as a payload-dropping mechanism and a unibody arm fixture, setting it apart from conventional commercial drones. This methodology prioritizes customization, enabling fine-tuning for specific applications, ensuring peak performance for its designated use-case.

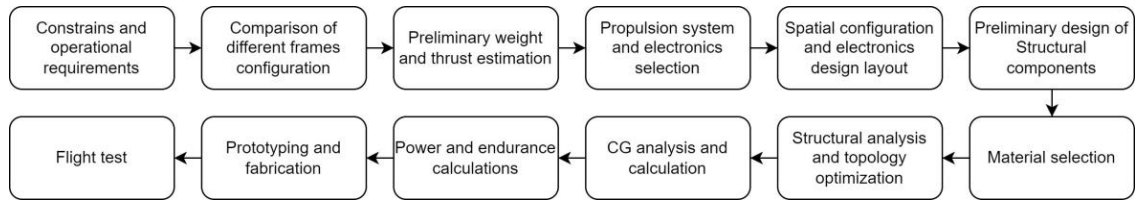

*Figure 1: Design Methodolgy*

## 3    DEMONSTRATION OF DESIGN METHODOLOGY

This section demonstrates the proposed methodology by designing and fabricating a drone adhering to the design constraints in subsection 3.1. Design methodologies employed and factors considered in each component and subsystem of a micro-UAV are emphasized.

### 3.1    Design Constraints

Establishing a comprehensive set of requirements and specifications is paramount before delving into the design and development of UAVs. The primary objective is to design a UAV specifically tailored for its payload applications. The subsequent table provides a detailed overview of the design's specifications and inherent limitations.

**Table 1: UAV Constraints**

| Sl. No | Parameter | Requirement/Limitation |
|--------|-----------|------------------------|
| 1. | UAV Category | Multirotor Micro UAS (Take-off weight < 2kg) |
| 2. | Payload Capacity | 200 grams |
| 3. | Payload Dimensions | 10 cm X 5 cm X 5 cm |
| 4. | Propulsion Type | Electric |
| 5. | Communication System Frequency | Data Telemetry – 2.4 GHz<br>Video Telemetry – 2.4 GHz or 5.8 GHz |
| 6. | Communication System Range | At least 1km |

### 3.2    Configuration Selection

Upon thoroughly examining various configurations, it was determined that a Quadcopter setup would be the most fitting choice to align with our mission-specific demands. While Hexacopters and Octocopters offer enhanced thrust capabilities, superior stability, and greater payload capacity, their inclusion of additional motors and associated accessories introduces supplementary weight, subsequently leading to heightened power consumption and, in turn, reduced flight duration. Conversely, a tricopter design presents a more simplistic and cost-effective alternative but has less controllability and payload capacity when compared to quadcopters. On the other hand, Co-axial drones stand out as high-performance UAVs reserved for missions with particularly stringent requirements.

### 3.3    Placement of Electronics:

Strategic placement of electronics is crucial in an UAV. After thoroughly researching the components needed and their specifications, a 3-layer arrangement of electronics and payload mechanism is designed based on the function and space constraints. The payload subsystem is placed at the bottom; all the electronics except the battery and Raspberry Pi are placed between the hub plates. As the battery requires sufficient airflow in its surroundings to ensure proper heat dissipation, it is placed on top of the hub plate and strapped using velcro straps. Due to its size, the Raspberry Pi is placed between the hub plate and the payload bay.

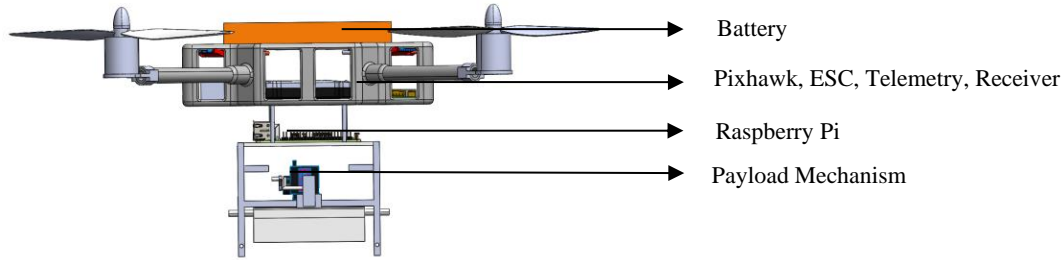

Battery

Pixhawk, ESC, Telemetry, Receiver

Raspberry Pi

Payload Mechanism

*Figure 3: Electronic Components Arrangement*

All electronic components were placed concisely to reduce the size, ease the circuital integration, and facilitate maintenance, ensuring easy accessibility for future repairs. Having this in mind, the preliminary shape of the hub design is optimized.

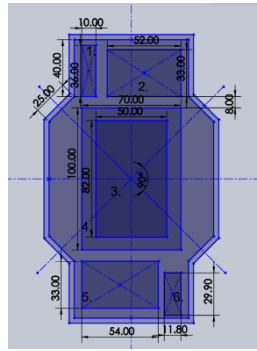

1. GPS Module: NEO 8M GPS with compass
2. Receiver: Flysky 6 channel
3. Flight controller: Pixhawk 2.4.8
4. Dampener
5. Telemetry -TS932 Video Rx/Tx
6. Servo - MG90s 13g
7. Motor- SunnySky X2212 1400kv

*Figure 4: Electronics Placement*

### 3.4    Conceptual Design

In a quadcopter, there are various frame configurations possible. The best-suited ones are Hybrid X, Hybrid H, and H frames, based on their simplicity and ability to accommodate all electronics. Preliminary design is prepared for three frame configurations and is subjected to a detailed comparative study.

#### 3.4.1    Hub Plate and Arm Fixture

The hub plate was designed in a simple octagon with extensions on the opposite sides to make it compact and aerodynamically efficient. The arm fixture was designed as the central part that connects all the components, providing significant support with a wider middle region, giving space to accommodate the wirings. It makes space for the extension to connect the arms. Specifically, the hub plate's design can be seamlessly integrated or modified to fit various frame configurations, including the Hybrid X, H, and H frames. This adaptability underscores the hub plate's role as a central, multifunctional component in the UAV's architecture.

**Table 2: Hub plate, Arm fixture and Assembly for various configurations**

| Frame configurations | Hybrid X | Hybrid H | H |
| --- | --- | --- | --- |
| Hub Plate | | | |
| Arm Fixture | | | |
| Hub Assembly | | | |

### 3.4.2 Arm and Motor Mount:

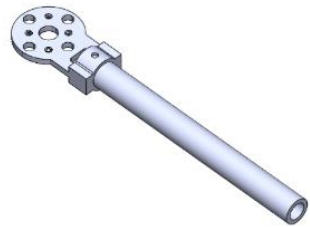

*Fig 4: Arm and motor mount*

A cylindrical cross-section carbon fibre tube of 12mm outer diameter and 10 mm inner diameter is chosen for the arm. Compared to a square tube, the cylindrical tube is far more effective in resisting torsional stress, has no stress concentration points, and distributes stress evenly, minimizing the chance of failure. The motor mount, manufactured using carbon fibre, integrates the motor and arms. It is designed with a plate-clamp model, in which the motor is placed on the flat plate, and the arm is fixed between the clamps, forming a robust and simple design that also provides effective cooling to motors.

### 3.4.3 Landing Gear

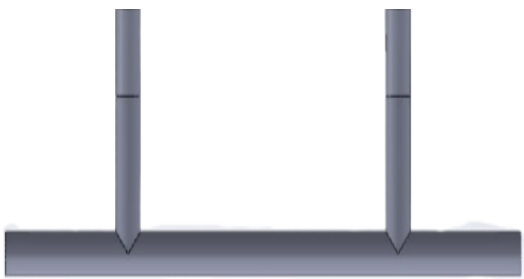

*Figure 5: Landing gear side view*

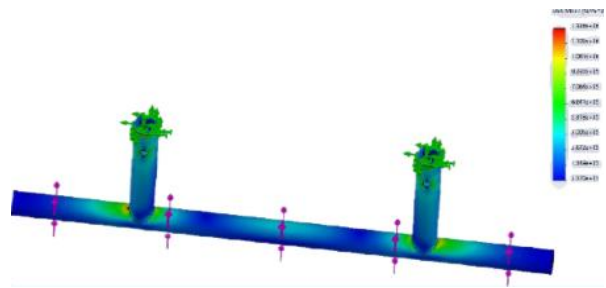

*Figure 6: FEM analysis of the landing gear*

The landing gear is supposed to withstand the freefall impact of the UAV and be stable enough to make the UAV stand upright during rough landing at improper angles. The base measures 0.21 m long, with a 0.01 m radius and 0.002 m thickness, with a 0.035 m protrusion from the center of the cylindrical base, inclined at 5°. The inclination provides a wider base for the UAV, thereby reducing the possibility of toppling. The total upward force experienced by the landing gear is considered to be twice the weight of the drone, which is close to 40 N. Structural analysis is done in SOLIDWORKS, considering the material to be ABS plastic with a tensile strength of $31.33 \times 10^6$ N/m² and the maximum stress exerted on the gear is $1.33 \times 10^6$ N/m².

### 3.4.4 Payload Bay and Dropping Mechanism

The payload considered is a solid cuboid of dimensions 0.1×0.05×0.05 m, weighing 0.2 Kg. The interior design of the payload bay was made to hold the payload tight enough with no room for movement. Ordinary dropping mechanisms like rolling or sliding doors always had the problem of high load on servos, hindering those mechanisms due to servos jamming. Commonly, this is mitigated by using high-torque servos, but they come at a cost of higher power consumption. In the mechanism designed, there is no load that affects the servos for their activation. Once activated, the torque required to open the latch reduces since the rod connects the servo and the latch. The spinning doors utilize gravity to drop the payload. The structural analysis of the latch is done in SOLIDWORKS FEM. The maximum stress exerted on the latch is $1.81 \times 10^6$ N/m² which gives a safety factor of 17.

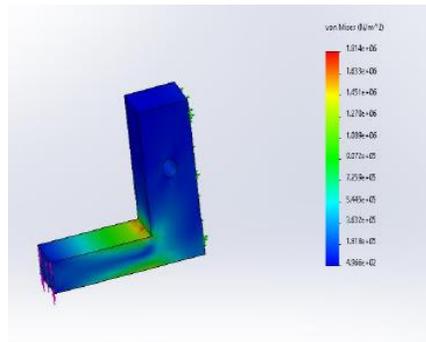
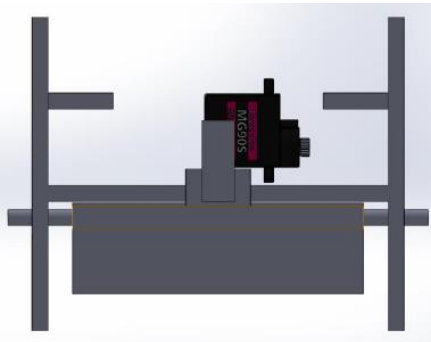
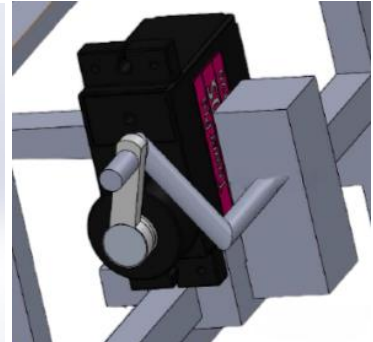

*Figure 7a: FEM analysis of latch*　　　*Figure 7b: Payload Bay Front View*　　　*Figure 7c: Locked Position*

## 3.5 Material Composition and Selection

This design incorporates the use of composite materials like Carbon-Fiber-Balsa sandwich and Carbon-Fiber Reinforced Polymer (CFRP). The hub plate consists of a lightweight balsa wood core enveloped by three-plies of carbon fiber and bound with epoxy resin. This unique configuration offers exceptional stiffness-to-weight and strength-to-weight ratios and ensures the drone's structural integrity and resistance to deformation.

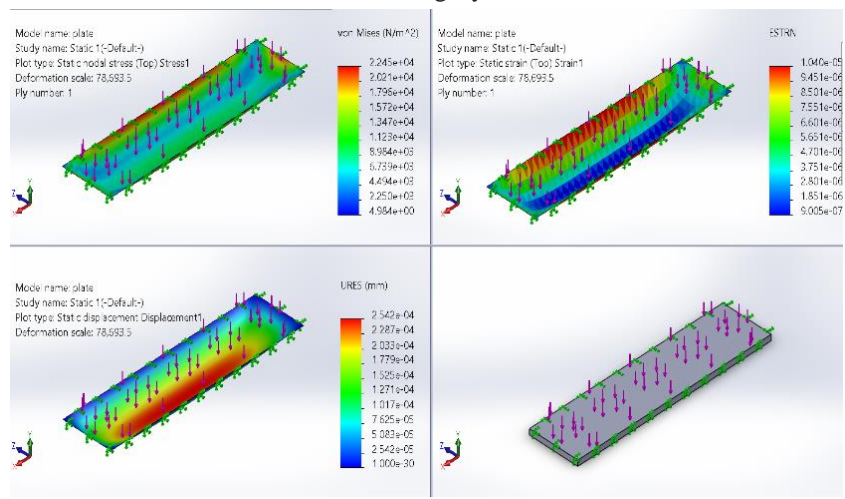

*Figure 8: Hub Plate Structural Analysis*

This effectively reduces the weight of the hub plate by 33% and cost by 42% comparing with a carbon-fiber plate with same thickness. When subjected to bending, the dynamics cause the bottom carbon fiber layer to stretch and the top to compress or vice-versa. This combination boasts a five times lower density and ten times higher tensile strength than conventional steel. With seven 0.00025 m thick piles, the carbon fiber laminate employs a quasi-isotropic layup, ensuring uniform strength and stiffness across all orientations. The 45°, 90°, and -45° ply orientation balance axial, shear, torsional, and transverse strengths. Various iterations of the structural analysis are done to finalize the resultant thickness. To ensure the carbon fiber laminate's optimal quality, the vacuum bagging process for fabrication is used, offering benefits such as reduced trapped air and optimized fiber-to-resin ratios.

To conclude, arm fixture is crafted from a 3D-printed chopped fiber-reinforced nylon composite selected for its tensile strength, lightweight nature, cost-effectiveness, and ease of fabrication. 3D printed Polylactic Acid (PLA) was used for the payload, landing gear and motor clamps. Carbon fiber was used for the arms and motor plate. The carbon-fiber balsa sandwiched composite hub plates, 3D-printed chopped fiber-reinforced composite arm fixture, and PLA payload bay and landing gear are held together using temporary fixtures such as threaded fasteners and spacers. The effect of differential thermal expansion would be crucial in the interaction between the hub plates and the arm fixture. However, it can be observed that the Coefficient of Thermal Expansion (CTE) for chopped-fiber reinforced polymer and the sandwiched composites are in the order of μm/°C, and the values are nearly the same. Furthermore, the operational temperature range depends on the environmental conditions, which do not vary drastically at a given location. Hence, the effect of differential CTE can be neglected.

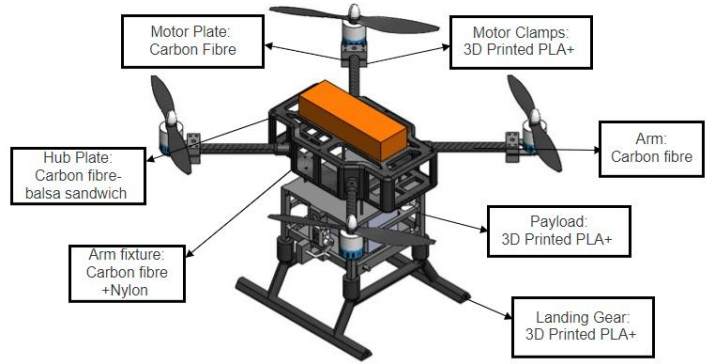

*Figure 9: Material Specification of Various Components*

## 3.6    Analysis and Optimization

### 3.6.1    Structural Analysis

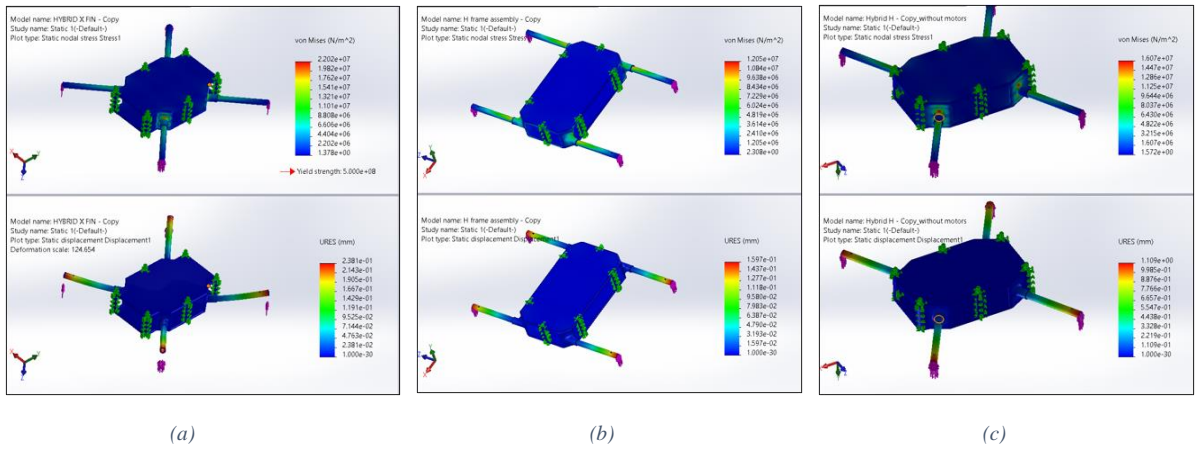

|        (a)        |        (b)        |        (c)        |

*Figure 10: Structural Analysis of (a)Hybrid X, (b)H, (c)Hybrid H frame configurations*

A structural analysis is conducted using SOLIDWORKS on three selected drone frame designs: Hybrid X, H, and Hybrid H. A load of 13 N is applied at the tips of the arms, determined as the product of the motor's maximum thrust and safety factor. The resulting deflections are $2.381 \times 10^{-4}$ m, $1.957 \times 10^{-4}$ m, and $1.109 \times 10^{-3}$ m for Hybrid X, H, and Hybrid H, respectively. The Von-Mises stress values at the joints between the arms and the body are $2.202 \times 10^{7}$ N/m², $1.2 \times 10^{7}$ N/m², and $1.607 \times 10^{7}$ N/m² for Hybrid X, H, and Hybrid H respectively. Upon comparing these results, it is evident that the Hybrid X drone frame exhibits superior structural performance when compared to the other two configurations. Consequently, the Hybrid X drone frame design is selected as the final choice due to its strength, stability, and economy compared to the H configuration.

### 3.6.2 Optimized Hub plate design

Considering the stress distribution, electronic placements, and fixtures to combine all the parts to the structure, top and the bottom hub plates are optimized, reducing the total weight of the preliminary hub plate design by ~30%.

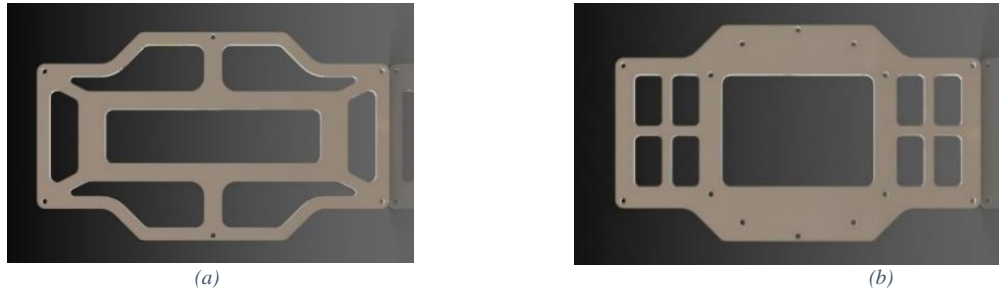

*(a)* *(b)*
*Figure 11: (a) Optimised Top plate,  (b) Optimised Bottom Plate*

### 3.6.3 C.G. and Stability Analysis

Centre of Gravity (CG) is obtained from the CAD model assembly with all the electronic components from SOLIDWORKS. An eigenvalue-based analysis is an integral part of our holistic UAV design approach. It enables us to assess the stability and maneuverability of the quadcopter, ensuring that it can operate reliably and effectively in various operational scenarios. The results of this analysis help us make subsequent design decisions and control algorithm tuning, contributing to the overall success of our autonomous UAV design.

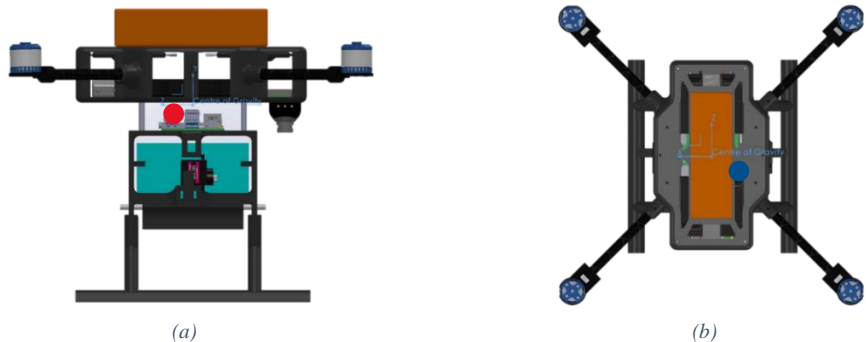

*(a)* *(b)*
*Figure 12: CG represented in Drone Assembly (a)Side and (b)Top views*

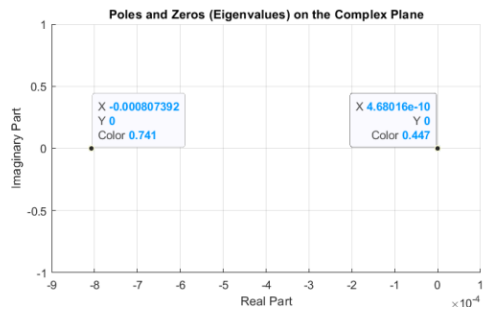

*Figure 13: Complex S-Plane (Pole Zero Plot)*

The entire stability analysis is performed with critical steps such as a) Drone dynamics, b) Control Inputs, and c) Compute the Matrix A and Visualization. Firstly, drone dynamics like mass and moment of inertia for accurate dynamic modelling is defined followed by control inputs, representing rates of motion (roll, pitch, yaw) and corresponding moments. Then the system matrix 'A' is computed, which represents the physical properties, control inputs and angular velocity change($\theta_c, \theta_y, \theta_z$) and moments($I_x, I_y, I_z$). The eigenvalues of matrix A are visualized in a plot which provides insights into system responses for disturbances and control effectiveness. Here, the real parts represent damping ratios, while imaginary parts indicate natural frequencies.

### 3.6.4 CFD Analysis

Computational Fluid Dynamics (CFD) analysis was conducted using ANSYS software to evaluate the drone's aerodynamic performance comprehensively. The CAD model of the drone, initially developed in SOLIDWORKS, was imported into ANSYS Design Modeler for further analysis. The propellers were meticulously meshed using the mesh sizing tool in ANSYS, with a maximum mesh size of 2.5 mm. This fine mesh resolution is vital for capturing the intricate flow patterns and turbulence near the propellers during operation. The quality of the mesh is essential to obtain accurate and reliable results.

**Table 3: Parameters for CFD Analysis**

| Parameter | Value |
|---|---|
| Propeller Rotation | 1300 rad/s |
| Gravity | -9.81 m/s² |
| Maximum Mesh Size | 2.5 mm |
| Viscous Model | k-epsilon |
| Time | Transient |

The virtual wind tunnel-type enclosure has been created around the Quadcopter. The parameters used for the CFD simulation are listed below. The number of iterations taken for the calculation is 200. The inlet velocity of air is 15 m/s. The static pressure at the outlet is 1 atm. From the results, the lift on the drone just above the propellers is about **19.65 N,** which is more than the estimated weight, and it experiences a drag force of **1.39291 N**.

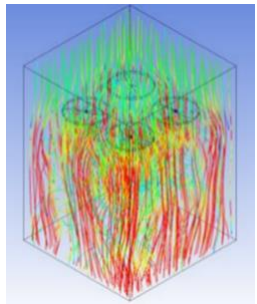 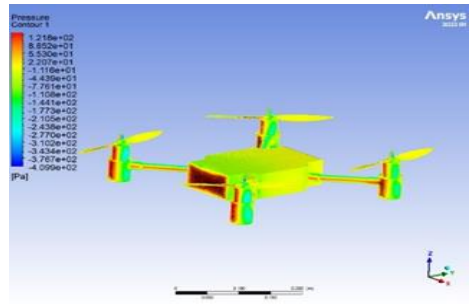

*Fig 16: Velocity Streamlines*          *Fig 17: Pressure Contours*

### 3.7 Autonomous system

The integration of autonomous systems in UAV for payload dropping with target detection represents a technological leap that enhances precision, adaptability, and safety in various applications.

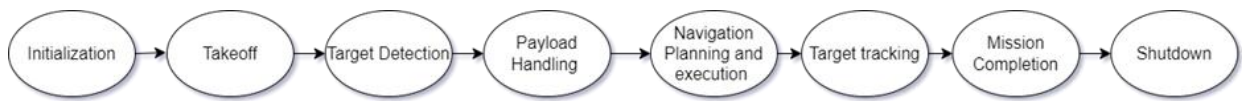

*Fig 19: Flow chart of overall autonomous operation*

To ensure the method is reliable and durable in a virtual environment, we have carefully simulated it using the Robot Operating System (ROS) framework and Gazebo simulation environment. Furthermore, we have made a significant effort with a two-phase strategy which includes both simulation and actual experimentation to ensure that the work is applicable in the real world by testing the algorithm's performance in real time.

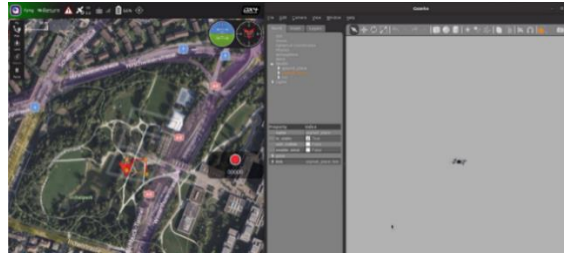

*Fig 20: SITL Simulation using Gazebo as Ground Station*

### 3.7.1 Navigation, Communication, and Object Tracking

The navigation system rested on the Mission Planner, harnessing the power of GPS for precise and robust navigation. This allows our UAV to autonomously plan and execute missions, ensuring that payload delivery to predefined target locations is achieved accurately. In tandem with this, we harnessed the advanced object detection capabilities of YOLO V5 and TensorFlow. This synergy of state-of-the-art computer vision tools enabled our UAV to detect and recognize specific target locations efficiently, ensuring the successful deployment of payloads. Object detection setup using SSD-Mobilenet and YOLO V5 was carried out sequentially. In our first approach, we implemented SSD (Single Shot Multi-Box Detector) with the MobileNet architecture. We used a training dataset of 150 labeled examples to train the SSD - MobileNet model. The training process involved optimizing the model's parameters to minimize the detection error. After training, the SSD - MobileNet model achieved an impressive accuracy rate of 99% on our test dataset.

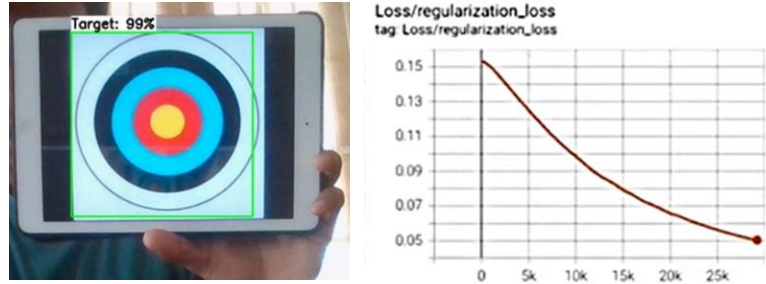

*Fig 21: Loss vs. training iterations curve*

**YOLOv5 Implementation:**

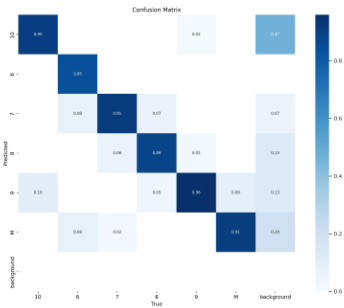

*Fig 22: Confusion Matrix*

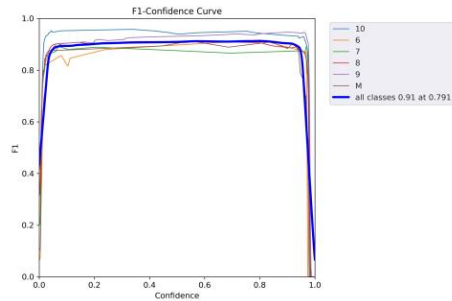

*Fig 23: F1 Vs Confidence curve*

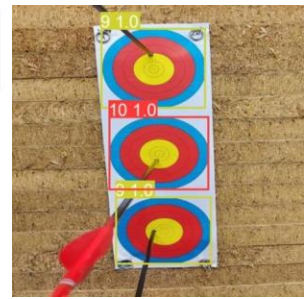

*Fig 24: Target recognition*

We implemented YOLOv5 in the second approach. We utilized a separate training dataset containing 1024 labeled examples for this framework. Training YOLOv5 involved fine-tuning the model's architecture and optimizing the associated hyperparameters. YOLOv5 demonstrated 90% accuracy rate, which was slightly lower than SSD-Mobilenet. Even though, computation time for YOLOv5 was higher, it proved to be more reliable.

**Comparison and Implications:**

The choice of using two different object detection frameworks allowed us to assess their respective strengths and weaknesses. The SSD-MobileNet model, with its extensive training dataset, demonstrated exceptional accuracy, making it suitable for high-precision scenarios. YOLOv5, despite a smaller training dataset, provided a good balance between accuracy and speed, making it a viable choice for real-time or resource-constrained applications. The choice between these frameworks ultimately depends on the specific requirements of the payload-dropping mission, such as the need for real-time performance or the degree of precision required.

### 3.8    Fabrication and Testing:

The hub plates in the drone play a critical role in housing essential electronic components. They are fabricated using a composite material made from carbon fiber and balsa wood. The carbon fiber plates are fabricated from 2.5 x $10^{-3}$ m thick carbon fiber sheets through a vacuum bagging technique, as illustrated below in the Fig.25. Vacuum bagging offers several advantages, including eliminating trapped air between layers, compression of fiber layers by efficient force transmission, and preventing shifting in fiber orientation while making a plate. The vacuum bagging process is critical for producing carbon fiber laminates, offering substantial benefits over alternative methods. It excels at enhancing consolidation, reducing voids, maintaining precise control over the fiber-to-resin ratio, and effectively reducing weight while improving surface quality and aesthetics. Moreover, this approach ensures a high level of consistency, allowing for the reliable and reproducible manufacture of composite components. Its versatility and scalability enable the production of customized, dependable components while minimizing waste and resource consumption, aligning with eco-friendly and sustainable manufacturing practices. Three layers of carbon fiber sheets are stacked with orientation perpendicularly to each other, one over the other on both the sides of the balsa core. An infusion mesh is placed over the carbon fiber sheets inside a bagging film to facilitate resin flow, which is forced through the plates using a vacuum pump. The entire setup is placed inside an oven for 18-22 hours at 60 degrees Celsius. Then, the bagging film and infusion mesh are removed to obtain the carbon fiber-balsa composite plate. Finally, the composite plates are machined to obtain the required shapes. The sandwich plates fabricated are also used to make the motor-mounts, thereby simplifying the manufacturing process while maintaining structural integrity. The arm fixture is fabricated by additively manufacturing Carbon-Fibre Reinforced Polymer (CFRP), Chopped Carbon-Fibre Nylon, which possesses high tensile strength while being extremely light. The payload bay and landing gear are additively manufactured using PLA material through 3D printing. The parts are finally assembled using fasteners. Post assembly, multiple testing of the drone was carried out successfully including autonomous payload dropping.

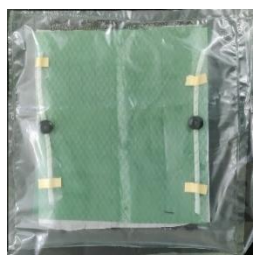

*Figure 25: Vacuum bag technique*

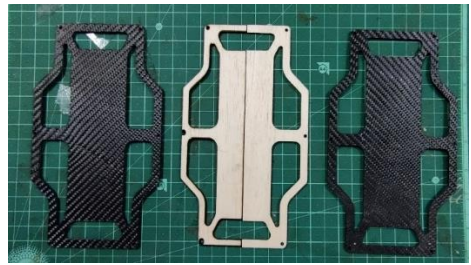

*Figure 26: CF, Balsa Layers*

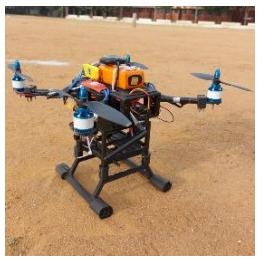

*Fig 27: Assembled UAV*

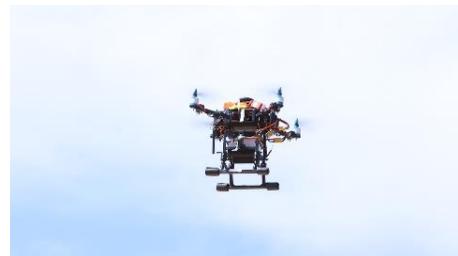

*Fig 28: UAV during flight*

## 4    CONCLUSION

The successful design and analysis of the multi-copter unmanned UAV have paved the way for diverse research opportunities in precise payload delivery. The entire process, from recognizing limitations to creating the prototype, has been effectively verified. Utilizing analytical tools aided in problem identification and design enhancement. The objective of autonomous drone control was attained through the application of machine learning. Recognizing the extensive potential of UAVs in the future and enhancing their design for optimal performance will present a challenge that can be surmounted by further refining the methodology, thereby bolstering efficiency and technological progress.

### ACKNOWLEDGEMENTS

We would like to express our sincere gratitude to the Third Dimension Aeromodelling Club, NIT Trichy for their generous funding and support, which made this research project possible. We extend our heartfelt thanks to Dr. R. Prakash from NIT Trichy for his invaluable assistance and for granting us access to the Composite labs for vacuum bagging, a crucial aspect of our project's success. We are also deeply appreciative of Dr. T. Ramesh from NIT Trichy for granting us access to the 3D printing facility at the CAD and CAM lab, NITT, which significantly contributed to our research endeavours. We are also grateful to RECAL for letting us use the SCIEnT facility for fabricating our drone. These contributions were instrumental in the successful completion of our research, and we are sincerely thankful for their support and guidance throughout the project.

### AUTHORSHIP CONTRIBUTION STATEMENT

**Aniruth A** : Conceptual Design, CAD, CFD, Payload Mechanism, CG and Stability Analysis, Fabrication, First Draft, Review and Editing; **Chirag Satpathy** : Initial Conceptual Design, Structural evaluation, Fabrication, Payload Mechanism, Composite Manufacturing, First Draft, Review and Editing; **Nitteesh M** : Autonomous, Control and Navigation Systems, Object Detection Models, Electronics and flight controller integration and calibration, CG and Stability Analysis, Fabrication, First Draft, Review and Editing; **Gokulraj M** : Initial Design(CAD), CFD, Fabrication, Autonomous Systems, Image Processing Techniques, First Draft, Review and Editing; **Jothika K** : Conceptual Design, CAD, CFD, Structural Analysis, Fabrication, First Draft, Review and Editing; **Venkatram K** : Payload Mechanism, Initial Design, CAD, Structural Analysis, Fabrication, First Draft; **Harshith G** : Initial Design, Arm-Fixture Design, First Draft; **Shristi S** : Initial Design, Fabrication, First Draft and Editing; **Anushka Vani** : Initial Design, Material Selection, Structural Analysis, Fabrication; **Jonathan Spurgeon**: CFD.

Autonomous UAV System Development for Payload Dropping Mission | Hadi | Journal of Instrumentation, Automation and Systems.